\pdfoutput=1

\documentclass[11pt]{article}

\usepackage{EMNLP2022}

\usepackage{times}
\usepackage{latexsym}

\usepackage[T1]{fontenc}

\usepackage[utf8]{inputenc}

\usepackage{microtype}

\usepackage{inconsolata}
\usepackage{booktabs}
\usepackage{makecell}
\usepackage{bbm}
\usepackage{graphicx}
\usepackage{amsmath}
\usepackage[footnotesize]{subfigure}

\title{Towards Robust $k$-Nearest-Neighbor Machine Translation}

\author{Hui Jiang\textsuperscript{1}\footnotemark[1],~~Ziyao Lu\textsuperscript{2}\footnotemark[1],~~Fandong Meng\textsuperscript{2},~~Chulun Zhou\textsuperscript{2},\\~~\textbf{Jie Zhou\textsuperscript{2},~~Degen Huang\textsuperscript{3}\and Jinsong Su\textsuperscript{1,4}\footnotemark[2]}\\
	\textsuperscript{1}School of Informatics, Xiamen University, China \\
	\textsuperscript{2}Pattern Recognition Center, WeChat AI, Tencent Inc, China\\
	\textsuperscript{3}Dalian University of Technology, China\\
	\textsuperscript{4}Pengcheng Laboratory, China\\
	\texttt{\small{hjiang@stu.xmu.edu.cn}} ~~~
	\texttt{\small{\{ziyaolu,fandongmeng,chulunzhou,withtomzhou\}@tencent.com}}\\
	\texttt{\small{huangdg@dlut.edu.cn}}~~~
	\texttt{\small{jssu@xmu.edu.cn}}
	\thanks{This work is done when Hui Jiang was interning at Pattern Recognition Center, WeChat AI, Tencent Inc, China.}
}
\makeatletter
\def\thanks#1{\protected@xdef\@thanks{\@thanks
		\protect\footnotetext{#1}}}
\makeatother
\begin{document}
	\maketitle
	\renewcommand{\thefootnote}{\fnsymbol{footnote}}
	\footnotetext[1]{Equal contribution}
	\footnotetext[2]{Corresponding author}
	\renewcommand{\thefootnote}{\arabic{footnote}}
	
\begin{abstract}
$k$-Nearest-Neighbor Machine Translation ($k$NN-MT) becomes an important research direction of NMT in recent years. 
Its main idea is to retrieve useful key-value pairs from an additional datastore to modify translations without updating the NMT model. 
However, the underlying retrieved noisy pairs will dramatically deteriorate the model performance.
In this paper,
we conduct a preliminary study and find that this problem  results from not fully exploiting the prediction of the NMT model.
To alleviate the impact of noise, we propose a confidence-enhanced  $k$NN-MT model with robust training.
Concretely, we introduce the NMT confidence to refine the modeling of two important components of $k$NN-MT: $k$NN distribution and the interpolation weight.
Meanwhile we inject two types of perturbations into the retrieved pairs for robust training.
Experimental results on four benchmark datasets demonstrate that our model not only achieves significant improvements over current $k$NN-MT models, but also exhibits better robustness. Our code is available at 
\url{https://github.com/DeepLearnXMU/Robust-knn-mt}.
\end{abstract}

\section{Introduction}
\begin{figure}[t]
	\centering
	\includegraphics[width=7.5cm]{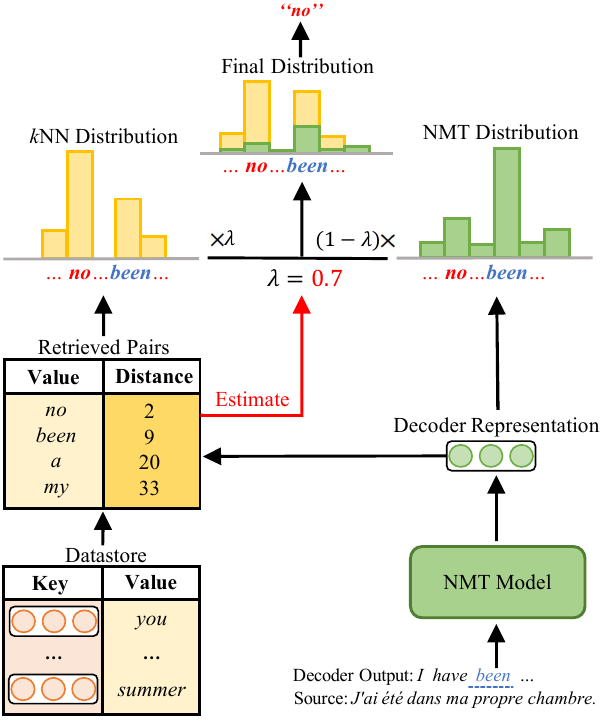}
	\caption{An example of $k$NN-MT model using dynamically estimation of $\lambda$ \cite{Zheng:ACL2021,Jiang:EMNLP2021}.}
	\label{fig1}
\end{figure}
As a commonly-used paradigm of retrieval-based neural machine translation (NMT), $k$-Nearest-Neighbor Machine Translation ($k$NN-MT) has proven to be effective in many studies \cite{Khandelwal:ICLR2021,Zheng:ACL2021,Jiang:EMNLP2021,Wang:ACL2022,Meng:ACL2022}, and thus attracted much attention in the community of machine translation.
The core of $k$NN-MT is to use an auxiliary datastore containing cached decoder representations and corresponding target tokens.
This datastore can  flexibly guide the NMT model to make better predictions, especially for domain adaptation. 
Compared with other retrieval-based paradigm \cite{Tu:TACL2018,Cai:ACL2021},
$k$NN-MT has two advantages:
1) It is more scalable because we can directly improve the NMT model by just manipulating the datastore.
2) It is more interpretable due to its observable retrieved pairs.

Generally, $k$NN-MT mainly involves two stages: datastore establishment
and candidate retrieval.
During the first stage, a pre-trained NMT model is used to construct a datastore containing key-value pairs, where the key is the 
decoder representation and the value is the corresponding target token.
At the second stage, given the current decoder representation as a query at each time step, the $k$ nearest key-value pairs are retrieved from the datastore. 
Then, according to the query-key
distances, the retrieved values are converted into a translation probability distribution over candidate target tokens, 
denoted as \emph{$k$NN distribution}.
Finally, the predicted distribution of the NMT model is interpolated by $k$NN distribution with a hyper-parameter $\lambda$.
Along this line, many efforts have been made to improve
$k$NN-MT \cite{Zheng:ACL2021,Jiang:EMNLP2021}. 
Particularly, as shown in Figure \ref{fig1}, adaptive $k$NN-MT \cite{Zheng:ACL2021} uses the query-key distances and retrieved pairs to dynamically estimate $\lambda$, 
exhibiting better performance than most $k$NN-MT models.

However, we find that existing $k$NN-MT models often suffer from a serious drawback: 
the model performance will dramatically deteriorate due to the underlying noise in retrieved pairs.
For example, in Figure \ref{fig1}, the retrieved results may contain unrelated tokens such as "no", which leads to a harmful $k$NN distribution.
Besides, for estimating $\lambda$, previous studies \cite{Zheng:ACL2021,Jiang:EMNLP2021} only consider the retrieved pairs while ignoring the NMT distribution.
Back to Figure \ref{1}, compared with the $k$NN distribution, the NMT model gives a much higher probability to the ground-truth token ``\emph{been}''.
Although the $k$NN distribution is insufficiently accurate, it is still assigned with a greater weight than the NMT distribution.
Obviously, this is inconsistent with our intuition that 
when the NMT model has high confidence in its prediction, it needs less help from others, and thus the $k$NN distribution should have a lower weight.
Moreover, we find that during training, a non-negligible proportion of the retrieved pairs from the datastore do not contain ground-truth tokens. This can cause insufficient training of $k$NN modules. To sum up, conventional $k$NN-MT models are vulnerable to noise in datastore, for which we further conduct a preliminary study to validate the above issues.
Therefore, dealing with the noise for the $k$NN-MT model remains to be a significant task.

In this paper, we explore a robust $k$NN-MT model.
In terms of model architecture, we explore how to more accurately estimate the $k$NN distribution and better combine it with the NMT distribution.
Concretely, unlike previous studies \cite{Zheng:ACL2021,Jiang:EMNLP2021} that only use retrieved pairs to dynamically estimate $\lambda$, we additionally use the confidence of NMT prediction to calibrate the calculation of $\lambda$ where confidence is the predicted probability on each retrieved token.
Meanwhile, we improve the $k$NN distribution by integrating the confidence to reduce the effect of noise.
Besides, we propose to boost the robustness of our model by randomly adding perturbations to retrieved key representations and augmenting retrieved pairs with pseudo ground-truth tokens.
By these means, our proposed approach can enhance the $k$NN-MT model to better cope with the noise in retrieved pairs,
thus improving its robustness.

To investigate the effectiveness and generality of our model, we conduct experiments on several commonly-used benchmarks. 
Experimental results show that our model significantly outperforms the adaptive $k$NN-MT, which is the state-of-the-art $k$NN-MT model, across  most domains.
Moreover, our model exhibits better performance than adaptive $k$NN-MT on pruned datastores.

\section{Related Work}
Retrieval-based approaches leveraging auxiliary sentences have shown effectiveness in improving NMT models.
Usually, they first retrieve relevant sentences from translation memory and then exploit them to boost NMT models during making a translation. For example,
\citet{Tu:TACL2018} maintains a continuous cache storing attention vectors as keys and decoder representations as values. The retrieved values are then used to update the decoder representations. \citet{Bapna:NAACL2019} preform n-gram retrieval to identify similar source n-grams from the translation memory, where the corresponding target words are then encoded to update decoder representations.  \citet{Xia:AAAI2019} pack the retrieved target sentences into a compact graph which is then incorporated into decoder representations. \citet{He:ACL2021} propose several Transformer-based encoding methods to vectorize retrieved target sentences. \citet{Cai:ACL2021} propose a cross-lingual memory retriever to leverage target-side monolingual translation memory, showing effectiveness in low-resource and domain adaption scenarios. 

Compared with the above studies involving additional training, non-parametric retrieval-augmented approaches \cite{Zhang:NAACL2018,Bult:ACL2019,Xu:ACL2020} are more flexible and thus attract much attention. According to word alignments, \citet{Zhang:NAACL2018} retrieve similar source sentences with target words from a translation memory, which are used to increase the probabilities of the collected target words to be translated. Both \citet{Bult:ACL2019} and \citet{Xu:ACL2020} retrieve related sentences via fuzzy matching and use the retrieved target sentences as the auxiliary information of the current source sentence.

Recently, a new non-parametric paradigm called $k$NN-MT \cite{Khandelwal:ICLR2021} has been proved to be simpler and more expressive. Typically, it uses the decoder representations as keys and the corresponding target words as values to build a datastore. 
During inference, based on retrieved results, the predicted distribution of the NMT model is interpolated by the $k$NN distribution with a hyper-parameter $\lambda$. 
Subsequently, some studies \cite{Zheng:ACL2021,Jiang:EMNLP2021} achieve better results by dynamically estimating $\lambda$. 
Meanwhile, there are also some researchers improving the retrieval efficiency of $k$NN-MT via cluster-based approaches \cite{Wang:ACL2022} or limiting the search space by source tokens \cite{Meng:ACL2022}. 
Besides, \citet{Zheng:EMNLP2021} presents a framework that uses in-domain monolingual target sentences to construct a datastore for unsupervised domain adaptation.

Finally, it should be noted that there have been many NLP studies \cite{Cheng:2018ACL,Cheng:2019ACL,Liu:2020AAAI,Miao:2022COLING} on exploring robustness of NLP models.
In comparison with the above-mentioned studies, 
our work is the first to improve the robustness of $k$NN-MT approaches.

\section{Preliminary Study}
\label{preliminary}
\begin{table}[t]
	\centering
	\begin{tabular}{c|l|c|c}
		\toprule
		\textbf{\#} & \textbf{Strategy}&  \textbf{\makecell[c]{Vanilla \\ $k$NN-MT}} & \textbf{\makecell[c]{ Adaptive \\ $k$NN-MT}} \\
		\midrule
		 1 & All & 45.92 & 47.88  \\
		 2 & ~-Random & 44.75 & 46.56  \\
		 3 & ~-$\left[0\%, 20\%\right)$    & 44.18 & 45.76 \\
		 4 & ~-$\left[20\%, 40\%\right)$    & 44.21 & 46.96 \\
		 5 & ~-$\left[40\%, 60\%\right)$    & 44.10 & 46.18   \\
		 6 & ~-$\left[60\%, 80\%\right)$    & 41.78 & 43.49  \\
		 7 & ~-$\left[80\%, 100\%\right]$    & 42.22 & 42.54 \\
		
		\bottomrule
    \end{tabular}
\caption{
	The performance of the models equipped with a datastore, where pairs
	within different intervals of ranking are individually removed.
	``-[0\%, 20\%)'' means the top 20\% pairs with the highest NMT confidence are removed.
	``All'' means the entire datastore is used, and ``-Random'' means 20\% pairs are randomly removed from the used datastore.
}
\label{table_pre}
\end{table}

\begin{figure}[t]
	\centering
	\includegraphics[width=7.5cm]{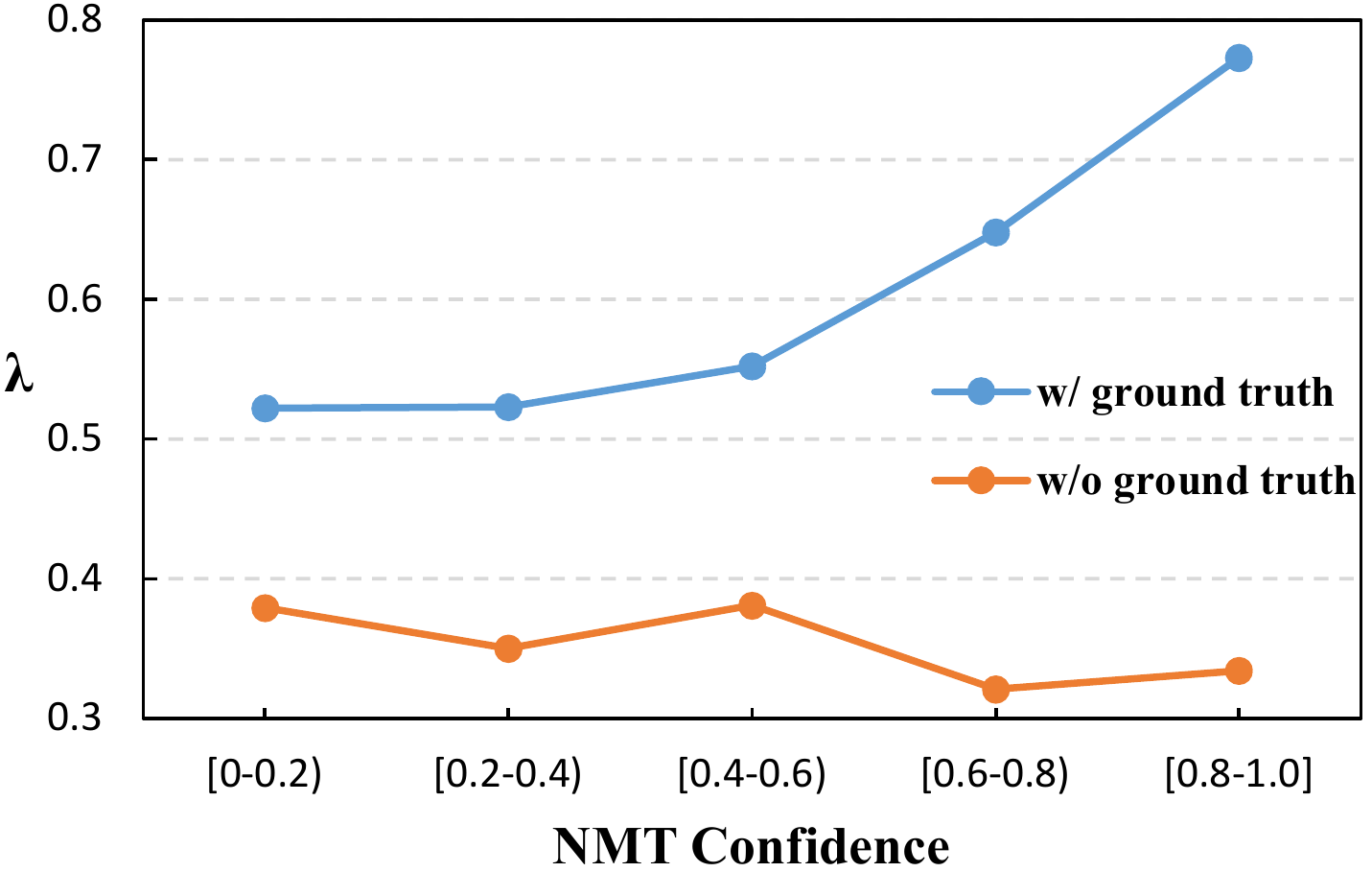}
	\caption{ $\lambda$ with respect to different NMT confidences on IT test set. Y-axis is $\lambda$, which is the weight of $k$NN distribution generated by adaptive $k$NN-MT. X-axis is the NMT confidence. We calculate the average generated $\lambda$ in different confidence intervals. The blue curve and orange curve represent the cases where the ground-truth token is retrieved or missed in the $k$NN distribution, respectively. }
	\label{fig_pre}
\end{figure}

To investigate the impact of noise on $k$NN-MT, we conduct a preliminary experiment in this section. 
We use the \emph{NMT confidence} to represent the predicted probability on the target token from the NMT model.
By this way,
we remove the pairs of datastore within different confidence intervals to investigate the impact of noisy datastore in different degrees of NMT confidence.

Specifically, during the datastore establishment, besides the key-value pairs, we additionally record 
the NMT confidence of each target token and use it to rank these pairs.
Then we split the datastore into multiple partitions,
we alternatively remove each partition of the datastore and observe the performance of vanilla $k$NN-MT \cite{Khandelwal:ICLR2021} and adaptive $k$NN-MT \cite{Zheng:ACL2021} on the IT training dataset \cite{Koehn:NMT2017}. To ensure the persuasiveness of our experiments, we directly use the setting of adaptive $k$NN-MT. In this way, we remove the key-value pairs within a specific interval of ranking to see the model performance. 

Table \ref{table_pre} lists the performance of the above two models. 
When removing the partition of the datastore within the interval [80\%, 100\%], we can observe that the performances of both models significantly degrade (See Row 7 in Table \ref{table_pre}), even inferior to ``-Random''.
It is reasonable because when the model has low confidence on its own prediction, 
it needs the retrieved pairs as supplementary information. 
Meanwhile, if we remove the high-confidence partition of datastore within the interval [0\%, 20\%), the performances of both models also decline (See Row 3 in Table \ref{table_pre}), underperforming the models with ``-Random''.
Intuitively, removing high-confidence partition should not have such a negative effect, as the NMT model is able to predict them correctly. 
We conjecture that this is because the retrieved pairs contain much noise after removing the high-confidence partition, harming the $k$NN distribution which is then used to interpolate the NMT distribution. 

Furthermore, since adaptive $k$NN-MT ignores NMT distribution, it may give a large weight to the incorrect $k$NN distribution. To verify this, we collect the $\lambda$ generated by adaptive $k$NN-MT with respect to different NMT confidences in Figure \ref{fig_pre}. 
Looking at the orange curve, we find that when the adaptive $k$NN-MT fails to retrieve the ground-truth token, it gives a similar weight $\lambda$ regardless of the NMT confidence. 
Besides, the blue curve shows the situation when the ground-truth token is successfully retrieved. We can see that adaptive $k$NN-MT gives a relatively small $\lambda$=0.52 even when the NMT model fails to predict the ground-truth (See [0-0.2) interval in Figure \ref{fig_pre}). Intuitively, the performance of the model would be further improved if it can generate a larger $\lambda$ when the model has unconfident NMT distribution and high-quality $k$NN distribution.

The above experimental results indicate that the $k$NN-MT models are sensitive to the quality of the datastore, which limits their applicability to a noisy datastore. Therefore, it is of great significance to explore robust $k$NN-MT.

\section{Our Model}
\begin{figure}[t]
	\centering
	\includegraphics[width=7.5cm]{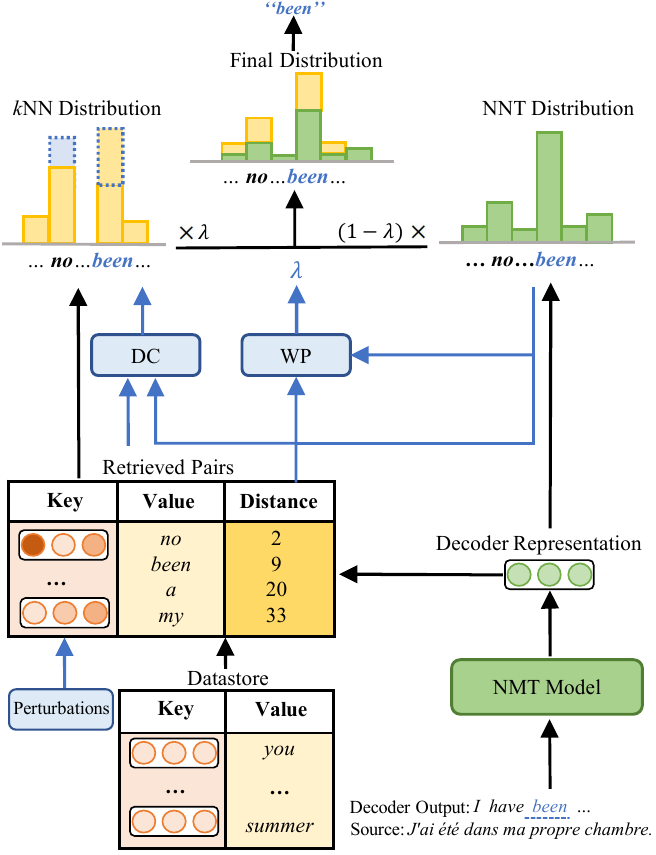}
	\caption{The overview of our confidence-enhanced $k$NN-MT model. The Distribution Calibration(DC) and Weight Prediction(WP) networks are trained to calibrate the kNN distribution and estimate the weight of $k$NN distribution, respectively.}
	\label{fig2}
\end{figure}

\subsection{Confidence-enhanced  $k$NN-MT}
Based on the observations in Section \ref{preliminary},
we can find that neglecting the prediction confidence of NMT model makes $k$NN-MT vulnerable to the noisy datastore.
Therefore, we leverage the prediction confidence of NMT model to enhance the robustness of $k$NN-MT, denoted as confidence-enhanced $k$NN-MT.
Similar to other $k$NN-MT models \cite{Khandelwal:ICLR2021,Zheng:ACL2021}, our model introduces a datastore to assist a pre-trained NMT model, involving two stages: datastore establishment and confidence-enhanced $k$NN-MT prediction.
Next, we give a full description of these two stages.

At the stage of datastore establishment, we adopt the pre-trained NMT model \cite{Vaswani:2017NIPS} to translate all training instances in an offline manner. During this process,
we record all decoder representations and their corresponding ground-truth target tokens as keys and values, respectively.
Formally, given a training set \{($x,y$)\}, we construct the datastore $D$ as follows:
\begin{equation}\label{1}
	\begin{aligned}
		D =  
		\{(h_t, y_t), \forall y_t \in y | (x, y) \},
	\end{aligned}
\end{equation}
where the key $h_t$ is the decoder representation of $y_t$, and the value $y_t$ is the corresponding ground-truth target token with $t$ denoting decoding timestep.

While inference, we firstly obtain the decoder representation $\hat{h}_t$ from the NMT model at the $t$-th timestep of decoding.
Afterwards, as implemented in the conventional $k$NN-MT \cite{Khandelwal:ICLR2021}, we  convert the retrieved pairs $N_t$=$\{(h_k,v_k), 1 $$\le$$ k $$ \le$$ K\}$ into a probability distribution over its values (i.e., $k$NN distribution).
Finally it is used to interpolate the NMT distribution to obtain a better translation.
Particularly, as shown in Figure \ref{fig2}, on the basis of previous $k$NN-MT models, we further introduce \textbf{D}istribution \textbf{C}alibration (\textbf{DC}) network and \textbf{W}eight \textbf{P}rediction  (\textbf{WP}) network, which leverage the model confidence to produce better $k$NN distribution and make more accurate estimation of $\lambda$, respectively.

Concretely,
we use the retrieved pairs $N_t$ and the decoder representation $\hat{h}_t$ to construct the $k$NN distribution.
Moreover, we propose the DC network to quantify the importance $c_k$ of each retrieved pair $(h_k, v_k)$, which is then used to refine the $k$NN distribution. Formally, the $k$NN distribution is constructed in the following way:
\begin{equation}\label{4}
	\begin{aligned}
		p_{\text{kNN}}(y_t | \hat{h}_t) \propto 
		\sum_{(h_k, v_k)\in N_t}\mathbbm{1}_{y_t=v_k}\text{exp}(\frac{-d_k}{T}+c_k),
	\end{aligned}
\end{equation}
\begin{equation}\label{eq-t}
	T = \mathbf{W}_{1}(\text{tanh}(\mathbf{W}_{2}[d_1,...,d_K;r_1,...,r_K])),
\end{equation}
\begin{equation}\label{3}
	c_{k} =  \mathbf{W}_{3}(\text{tanh}(\mathbf{W}_{4}[p_{\text{NMT}}(v_k|\hat{h}_t);p_{\text{NMT}}(v_k|h_k)])),
\end{equation}
where $d_k$ is the $L_2$ distance between query $\hat{h}_t$ and key $h_k$, $r_k$ is the number of non-duplicate values in top $k$ neighbors,
and $\mathbf{W}_*$ are parameter matrices.\footnote{In this paper, all $\mathbf{W}_*$ denote parameter matrices.}
Here, when calculating $c_k$, we mainly consider two kinds of information:
1) $p_{\text{NMT}}(v_k|\hat{h}_t)$, the predicted probability on $v_k$ from the NMT model given the decoder representation $\hat{h}_t$, and
2) $p_{\text{NMT}}(v_k|h_k)$, the predicted probability on $v_k$ given the key $h_k$.\footnote{We use the logarithm of probability as the feature, and we simplify the formula by omitting the log in this paper.}
In this way, the $k$NN distribution can be optimized by exploiting the knowledge of NMT, where the pairs with low confidence are expected to be assigned with lower probabilities.

However, this still can not make the model sufficiently robust to the noisy datastore.
As mentioned previously in the introduction and preliminary study, when 
the retrieved pairs contain much noise, it is not appropriate to estimate $\lambda$ only based on retrieved pairs \cite{Zheng:ACL2021}.
Therefore, we propose a lightweight WP network that simultaneously exploits the confidence of $k$NN distribution and NMT distribution to dynamically estimate $\lambda_t$:
\begin{equation}\label{6}
	\lambda_t = \frac{\text{exp}(s_{k\text{NN}})}{\text{exp}(s_{k\text{NN}}) + \text{exp}(s_{\text{NMT}})},
\end{equation}
\begin{equation}\label{7}
	s_{k\text{NN}} = \mathbf{W}_{5}(\text{tanh}(\mathbf{W}_{2}[d_1,...,d_K;r_1,...,r_K])),
\end{equation}
\begin{equation}\label{8}
	\begin{aligned}
	s_{\text{NMT}} = \mathbf{W}_{6}&[p_{\text{NMT}}(v_1|\hat{h}_t),...,p_{\text{NMT}}(v_K|\hat{h}_t); \\
	&p_{\text{NMT}}(v_1|h_1),...,p_{\text{NMT}}(v_K|h_K);\\
	&p_\text{NMT}^{top1},...,p_\text{NMT}^{topK}],
\end{aligned}
\end{equation}
where $p_\text{NMT}^{topk}$ is the $k$-th highest probability of the NMT distribution.

Lastly, the final distribution is computed as 
\begin{equation}\label{5}
	p(y_t | x, y_{<t}) = \lambda_{t} p_{\text{kNN}} + (1- \lambda_t) p_{\text{NMT}}.
\end{equation}
By doing so, we expect that $p_{\text{kNN}}$ will be assigned with a small $\lambda_t$ if the NMT model is highly confident on the predicted token. 

\begin{table*}[t]
	\centering
	\setlength{\tabcolsep}{1mm}
	\begin{tabular}{l|cccc|c}
		\toprule
		\textbf{Model} &  \textbf{IT}&  \textbf{Medical}&  \textbf{Koran}&  \textbf{Law}&  \textbf{Avg.}\\
		\midrule
		base NMT & 38.35 / 0.391 & 40.06 / 0.468 & 16.26 / -0.018 & 45.48 / 0.574 & 35.04 / 0.354  \\
		vanilla $k$NN-MT & 45.92 / 0.531 & 54.46 / 0.548 & 20.29 / -0.014 & 61.27 / 0.662  &  45.48 / 0.432\\
		adaptive $k$NN-MT & 47.88 / 0.567 & 56.10 / 0.572 & 20.43 / 0.037 & 63.20 / 0.692 &  46.90 / 0.467 \\
		\midrule
		our model & \textbf{48.90}$\ddagger$ / \textbf{0.585}$\ddagger$ & \textbf{57.28}$\ddagger$ / \textbf{0.578} & \textbf{20.71} / \textbf{0.047}$\dagger$ & \textbf{64.07}$\ddagger$ / \textbf{0.703}$\ddagger$ & \textbf{47.74} / \textbf{0.478} \\
		
		\bottomrule
	\end{tabular}
	\caption{
		The BLEU (\%) / Comet scores on test sets of different domains.
		$\dagger$ or $\ddagger$: significantly better than adaptive $k$NN-MT with t-test p<0.05 or p<0.01. Here we conducted 1,000 bootstrap tests \cite{Koehn:2004EMNLP} to measure the significance in score differences.
	}
	\label{main_result}
\end{table*}
\subsection{Model Training}
\begin{figure}[t]
\centering
\vspace{0.2cm}
\subfigure[Adding noise to the retrieved keys.]{
	\captionsetup{font={tiny}}
	\label{fig:subfig:1} 
	\includegraphics[width=7.5cm]{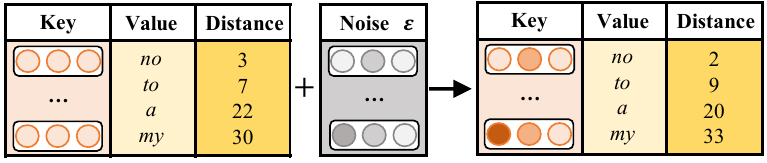}}
\subfigure[Constructing a pseudo pair with the ground-truth token as value. The pseudo pair in the blue dotted box is inserted into the retrieved pairs and the pair in the red dotted box is removed.]{
	\label{fig:subfig:2} 
	\includegraphics[width=7.5cm]{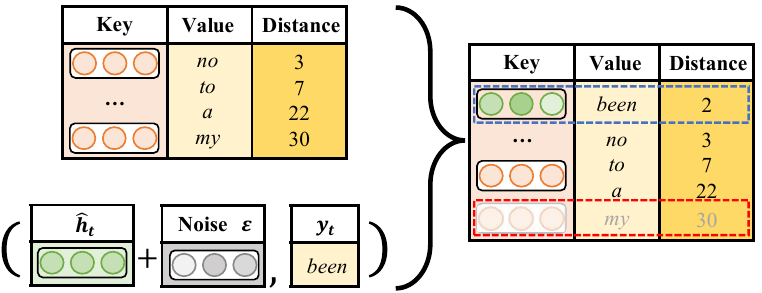}}
\caption{Two different perturbations are used for robust training. In both cases, the ground-truth token "been" is not retrieved by the $k$NN-MT model.}
\label{fig3} 
\end{figure}
Although through the above modification, our model is able to generate a better $k$NN distribution and make a more accurate estimation of $\lambda$, it may be still not robust enough for two reasons. 
First, the datastore may be incompatible with the test set, resulting in the retrieved pairs cannot help the model.
Second, the retrieved pairs do not always contain the ground-truth token.
In that case,
the probability of this token is zero in the $k$NN distribution.
As a result, 
our DC network will not be optimized on this training sample. 
Especially, when the datastore size is limited, these two problems are more serious. To address them, we  
add two types of perturbations to the retrieved pairs at the training stage.

For the first problem, as shown in Figure \ref{fig:subfig:1},  
we add Gaussian noise to the keys of retrieved pairs with a certain ratio $\alpha$.
At each training timestep, we generate a random value between 0 and 1 and add noise only if it is less than $\alpha$, so as to construct a noisy datastore:
\begin{equation}\label{9}
	h^{\prime}_k =  h_k + \mathbf{\epsilon}, ~~\mathbf{\epsilon \sim N(0,\sigma^2 \mathbf{I})},
\end{equation}
where the noise vector $\epsilon$ is sampled from a Gaussian distribution with variance
$\sigma^2$, and $\sigma$ is set to 0.01  as implemented in \citet{Cheng:2018ACL}. 

For the second problem, as shown in Figure \ref{fig:subfig:2}, we construct pseudo retrieved pairs with ground-truth tokens as values.
Specifically, at the $t$-th timestep,
if the ground-truth token $y_t$ is not retrieved, we use the current decoder representation $\hat{h}_t$ and $y_t$ to construct a pseudo pair $(\hat{h}_t + \mathbf{\epsilon}, y_t)$.
Then, we add this pair into the retrieved pairs $N_t$, where the pairs are sorted according to query-key distances, and the pair with the largest distance is removed to ensure the pair number is unchanged.
Similarly, this perturbation vector $\mathbf{\epsilon}$ is added with the same ratio $\alpha$.

However, we find that using a fixed perturbation ratio results in performance degradation. 
We speculate that applying too large perturbations in the final training stage impairs the model's ability to handle real samples in the datastore. To avoid its negative impact,
we dynamically adjust the perturbation ratio $\alpha$ according to the training step:
\begin{equation}\label{noise_rate}
	\alpha = \alpha_0 * \text{exp}(-step/\beta),
\end{equation}
where $\alpha_0$ and $\beta$ control the initial value and the declining speed of $\alpha$, respectively. By doing so, we expect the perturbation ratio to be large at the beginning and gradually decrease during the subsequent stages.

\section{Experiments}
To investigate the effectiveness and robustness
of our model, we carry out experiments on several
commonly-used datasets.

\subsection{Experimental Settings}
\subsubsection{Datasets and Evaluation}
\begin{table}[t]
	\centering
	\setlength{\tabcolsep}{2mm}
	\begin{tabular}{c|c|c|c|c}
		\toprule
		\textbf{Dataset} &  \textbf{IT}&  \textbf{Medical}&  \textbf{Koran}&  \textbf{Law}\\
		\midrule
		Train & 223K & 248K & 18K & 467K  \\
		Dev & 2K & 2K & 2K & 2K   \\
		Test & 2K & 2K & 2K & 2K  \\
		\midrule
		\makecell[c]{Size} & 3.71M & 6.90M & 524K & 19.0M \\
		
		\bottomrule
	\end{tabular}
	\caption{
		The statistics of datasets in different domains.
		We also list the size of the datastore, which is the number of stored tokens.
	}
	\label{dataset}
\end{table}
To ensure fair comparisons, we follow \citet{Zheng:ACL2021} to conduct  experiments on four commonly-used benchmarks, of which domains include IT, Koran, Medical and Law. 
The details of these datasets are given in Table \ref{dataset}.
We use the Moses toolkit\footnote{https://github.com/moses-smt/mosesdecoder} to tokenize sentences and split words into subword units \cite{Sennrich:ACL2016}.
As for the datastore, we adopt Faiss \cite{Johnson:TBD2021} to conduct quantization and retrieval.
Finally, all translation results are
evaluated with case-sensitive detokenized BLEU by SacreBLEU \cite{Post:WMT2018},
we also adopt the Comet \cite{Rei:2020EMNLP} as a complementary metric.

\subsubsection{Baselines}
We use the following models as our baselines:

\begin{enumerate}
	\setlength{\itemsep}{0pt} 
	\setlength{\parsep}{0pt} 
	\setlength{\parskip}{0pt} 
	\item \textbf{Base NMT}. We use the winner model \cite{Ng:2019WMT} of WMT'19 German-English news translation task as the base NMT model, which is also used to initialize other $k$NN-MT models.
	\item \textbf{Vanilla $k$NN-MT} \cite{Khandelwal:ICLR2021}. It is our basic baseline. Note that it tunes hyper-parameters including $\lambda$ on development sets.
	\item \textbf{Adaptive $k$NN-MT} \cite{Zheng:ACL2021}. It is our most important contrastive model that uses a light-weight network to dynamically estimate $\lambda$.
\end{enumerate}

As for our model, we empirically set the hidden size of our WP and DC networks to 4 and 32, respectively, and the number of retrieved pairs ($K$) is set to 8 in all experiments. We empirically set $\alpha_0$ to 1.0 and $\beta$ to 1000, except for the Koran dataset where $\beta$ is set to 10 due to its small data size.
During the model training, we use the development sets to train our networks for about 5K steps following \citet{Zheng:ACL2021}. As for other hyper-parameters, we use the same experimental setup as adaptive $k$NN-MT, so as to ensure fair comparisons. We use Adam to optimize our networks, the batch size is set to 32, and the learning rate is set to 3e-4.
All experiments are conducted on one NVIDIA V100 GPU.

\subsection{Main Results}
Table \ref{main_result} shows the main results.
Echoing previous studies \cite{Khandelwal:ICLR2021,Zheng:ACL2021},
vanilla $k$NN-MT significantly outperforms base NMT on all datasets.
Moreover, due to the advantage of dynamic $\lambda$, adaptive $k$NN-MT exhibits significant improvement compared to vanilla $k$NN-MT on most datasets except for Koran, where only 18K training samples are available. Furthermore, our model achieves the best performance, obtaining the average +0.84 BLEU score over adaptive  $k$NN-MT, which demonstrates the effectiveness of our model and training strategy.
This conclusion remains valid when testing with the Comet score, where our model still outperforms adaptive $k$NN-MT.

Note that on the Koran dataset, adaptive $k$NN-MT only performs slightly better than vanilla $k$NN-MT. Likewise, our model achieves a slight improvement over adaptive $k$NN-MT. For these results, we speculate that the extremely small size of the datastore for Koran limits the potential of both adaptive $k$NN-MT and our model.

\subsection{Robustness of Our Model}
\begin{table}[t]
	\centering
	\begin{tabular}{c|cc|c}
		\toprule
		\textbf{\makecell[c]{Pruning \\ Rate}} &  \textbf{\makecell[c]{Adaptive \\ $k$NN-MT}} & \textbf{\makecell[c]{Our model}} & \textbf{Size}\\
		\midrule
		0\% & 47.88 & 48.90 & 3.6M\\
		20\% & 46.56 & 47.30 & 2.9M\\
		40\% & 44.44 & 44.95 & 2.2M\\
		60\% & 42.24 & 42.69 & 1.4M\\
		80\% & 39.87 & 40.22 & 0.7M\\
		
		\bottomrule
	\end{tabular}
	\caption{
	The BLEU scores of the models equipped with the randomly reduced datastores on IT dataset.
	}
	\label{robust1}
\end{table}
\begin{table}[t]
	\centering
	\begin{tabular}{c|cc|c}
		\toprule
		\textbf{\makecell[c]{Pruning \\ Rate}} &  \textbf{\makecell[c]{Adaptive \\ $k$NN-MT}} & \textbf{\makecell[c]{Our model}} & \textbf{Size}\\
		\midrule
		0\% & 47.88 & 48.90 & 3.6M\\
		20\% & 45.76 & 47.68 & 2.9M\\
		40\% & 42.43  & 46.89 & 2.2M\\
		60\% & 41.05  & 45.52 & 1.4M\\
		80\% & 38.79  & 41.02 & 0.7M\\
		
		\bottomrule
	\end{tabular}
	\caption{
	The BLEU scores of the models equipped the reduced datastores on IT dataset, where the pairs having top x\% largest NMT confidence are removed.
	}
	\label{robust2}
\end{table}

To verify the robustness of our model,
we explore the performance of models with retrieved pairs of different qualities.  Specifically, we decrease the quality of retrieved pairs by pruning the datastore in the following two ways and then test our model.

Firstly, we randomly remove the pairs of datastore and report the performance of our model and adaptive $k$NN-MT in Table \ref{robust1}. 
Overall, our model performs well in all situations and even surpasses adaptive $k$NN-MT by 0.35 BLEU when reducing the size of the datastore to 20\%.
It is reasonable to observe that the performances of two models get closer when the size of datastore becomes smaller.

Secondly, we conduct another experiment on datastore pruning from the perspective of NMT confidence. 
Intuitively, words with higher NMT confidence are less necessary to be saved as they are easier to be correctly predicted by the NMT model. Thus, we rank all datastore pairs according to their NMT confidence and remove those with the largest NMT confidence. Table \ref{robust2} reports the experimental results.
Compared to adaptive $k$NN-MT, our model exhibits much less performance decline.
Particularly, when the datastore is compressed to 40\%, our model still outperforms adaptive $k$NN-MT by a large margin (+ 4.47 BLEU). 
This result demonstrates the potential of our model on pruned datastores.

\subsection{Analysis}
We also study the effect of the important hyper-parameters: the number of retrieved pairs ($K$), to further validate the robustness of our model.

	\begin{figure}[t]
		\centering
		\includegraphics[width=7.5cm]{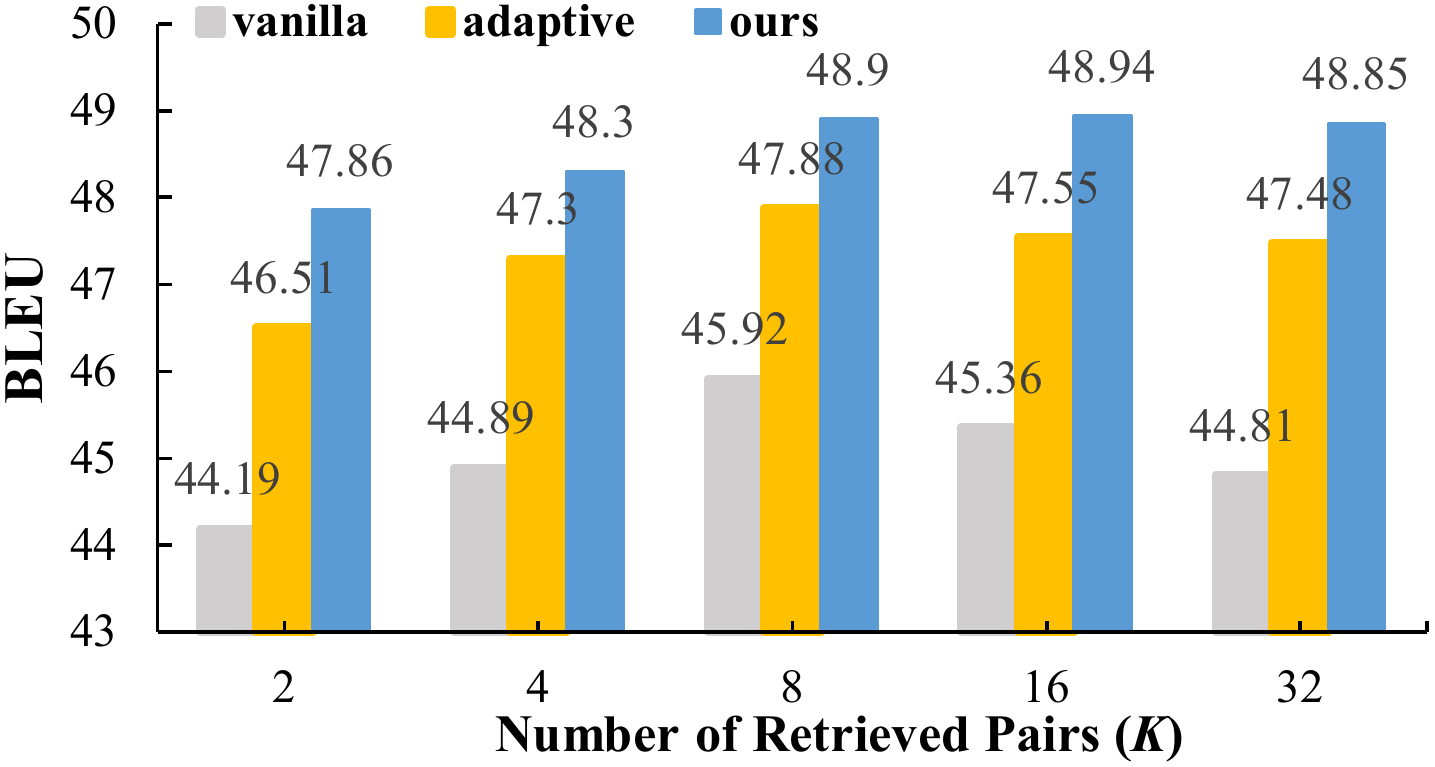}
		\caption{			The BLEU scores of the models equipped with different $K$ on IT dataset.}
		\label{analyse_k}
	\end{figure}

As shown in Figure \ref{analyse_k}, we find that performance of both vanilla $k$NN-MT and adaptive $k$NN-MT are not further improved when increasing K. This is because retrieving more neighbors may add noise to the $k$NN distribution. However, our model has a better performance when $K$=16. Overall, our model always exhibits better performance than adaptive $k$NN-MT especially when $K$ is large, demonstrating its robustness.

%

\begin{figure}[t]
	\centering
	\includegraphics[width=7.5cm]{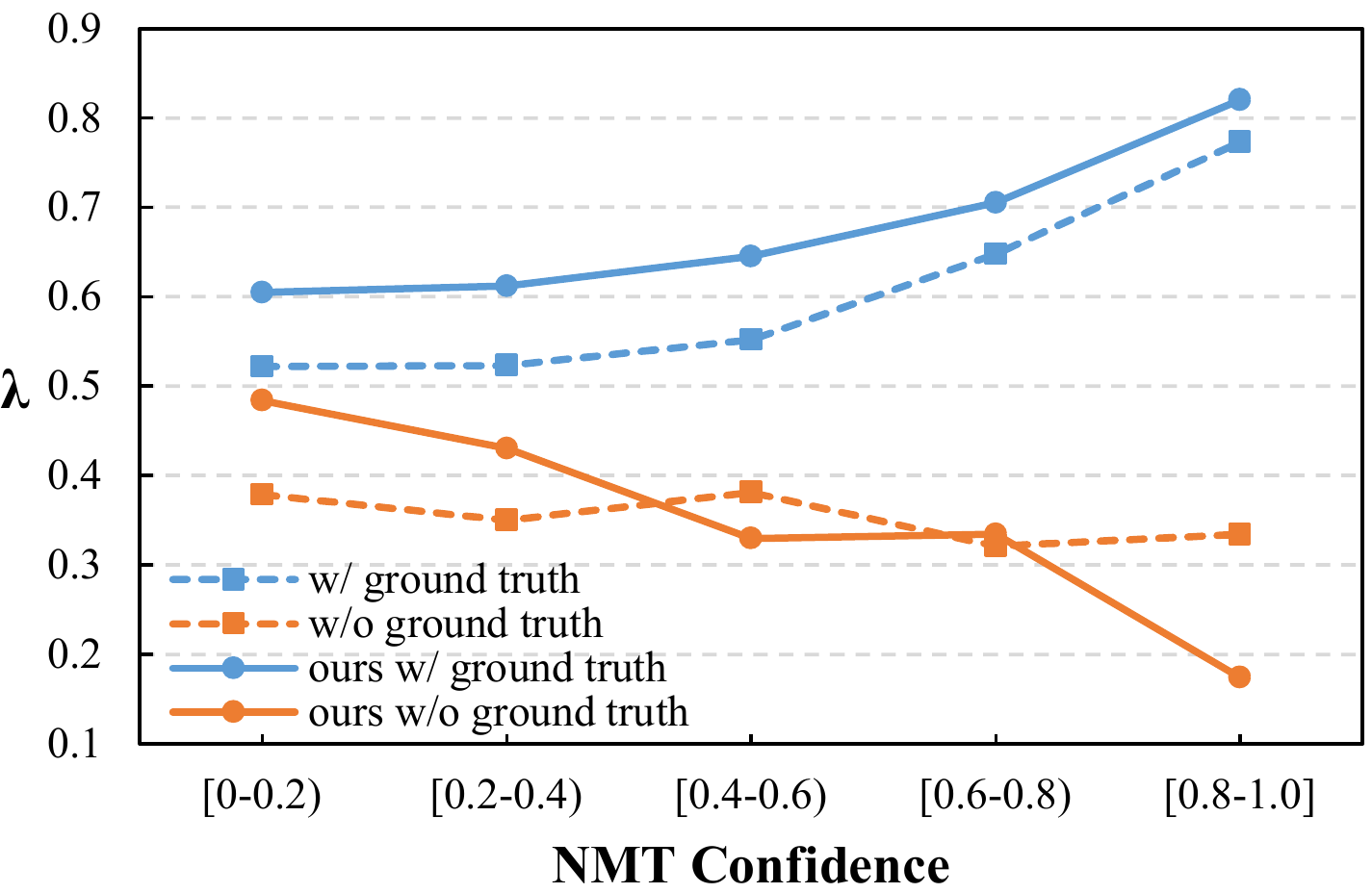}
	\caption{ $\lambda$ with respect to different NMT confidences on IT test set. We calculate the average generated $\lambda$ in different confidence intervals. Solid and dotted lines represent that $\lambda$ is generated by our model or adaptive $k$NN-MT, respectively. }
	\label{analy_lambda}
\end{figure}

Besides, to verify that our model is able to generate a better $\lambda$ to improve the final translation,
we study the predicted $\lambda$ within different confidence intervals.
Figure \ref{analy_lambda} reports the experimental results on the IT test set.
The blue curve represents the situation when the ground-truth token is successfully retrieved. We can see that the $\lambda$ generated by our model is larger than that generated by adaptive $k$NN-MT, especially when the NMT model fails to predict the ground truth (See [0-0.2) interval in Figure \ref{analy_lambda}). 
Looking at the orange curve, we find that when models fail to retrieve the ground-truth token, the generated $\lambda$ by our model has a stronger correlation with the NMT confidence. When model has confident NMT distribution (See [0.8-1.0] interval in Figure \ref{analy_lambda}),
our model can generate a lower weight ($\lambda$) of $k$NN distribution than adaptive $k$NN-MT.

Overall, it shows that our model can dynamically estimate the $\lambda$ based on the NMT confidence. It also confirms our assumption mentioned in
the preliminary study that the high-confidence prediction NMT distribution is expected to be assigned with a greater $\lambda$.

\subsection{Ablation Study}
\begin{table}[t]
	\centering
	\setlength{\tabcolsep}{5mm}
	\begin{tabular}{l|c}
		\toprule
		\textbf{Model} &  \textbf{BLEU}\\
		\midrule
		vanilla $k$NN-MT & 45.92 \\
		adaptive $k$NN-MT & 47.88\\
		\midrule
		our model & \textbf{48.90} \\
		~~w/o WP network & 48.36 \\
		~~w/o DC network & 48.45 \\
		\midrule
		~~w/o vector perturbation &  48.77 \\
		~~w/o pseudo pair perturbation &  48.75 \\
		~~w/o robust training & 48.68 \\
		~~w/o perturbation rate's decline  & 48.38 \\
		
		\bottomrule
	\end{tabular}
	\caption{
    Ablation study of different networks and training strategies on IT test set. ``w/o robust training'' means removing both the ``vector perturbation'' and ``pseudo pair perturbation'' training strategy.
	}
	\label{ablation}
\end{table}
To investigate the effects of our proposed networks and training strategies on our model,
we also provide the performance of different variants of our model.
As shown in Table \ref{ablation}, 
we find that removing any proposed network or not using any training strategy leads to a performance decline, 
demonstrating the effectiveness of all proposed networks and training strategies.
Particularly, when discarding the WP network for prediction of $\lambda$, our model shows the most significant performance drop.

As for our training strategy, ``w/o vector perturbation'' represents removing the perturbation of the key vector (See Equation \ref{9}),
``w/o pseudo pair perturbation'' means removing the perturbation of constructing pseudo pair.
It shows that constructing pseudo pair is more effective.
It should be noted that if we do not decrease the perturbation rate, the model performance will degrade severely because of the overwhelming noise.

\section{Conclusion}
In this paper, via preliminary study, we first point out that existing $k$NN-MT models are very susceptible to the quality of retrieved pairs. Then, we explore robust $k$NN-MT, which improves $k$NN-MT models in the aspects of model architecture and training. 
Concretely, we incorporate the confidence of NMT prediction into modeling $k$NN distribution and dynamic estimation of $\lambda$.
Besides, during the model training, we inject two types of perturbations into the retrieved pairs, which can effectively enhance the generalization of the model. Extensive results and in-depth analysis strongly demonstrate the effectiveness  of our model. 

To further verify the generality of our model, we will extend our model to other conditional text generation tasks, such as speech translation.
Besides, we will try to combine $k$NN-MT 
with topic information, which has been successfully applied in previous studies \cite{Su:2012ACL,Yu:2013IJCNLP,Su:2015ACL,Ruan:2018PRICAI}, to constraint retrieval in the future.

\section*{Limitations}
In terms of efficiency, the storage cost of datastore and the time cost of retrieval are proportional to the size of training data and thus quite high for $k$NN-MT models. 
Besides, our model involves an additional small amount of parameters compared to vanilla $k$NN-MT \cite{Khandelwal:ICLR2021}, requiring at least some in-domain samples for training. Although it can be applied in low-resource scenarios, it is not suitable for the scenario where in-domain samples are extremely few.

\section*{Acknowledgements}
The project was supported by  
National Key Research and Development Program of China (No. 2020AAA0108004), 
National Natural Science Foundation of China (No. 62276219), 
Natural Science Foundation of Fujian Province of China (No. 2020J06001),
Youth Innovation Fund of Xiamen (No. 3502Z20206059),
and the Major Key Project of PCL.
We also thank the reviewers for their insightful comments.

\bibliography{anthology,custom}
\bibliographystyle{acl_natbib}

\appendix

%

\end{document}